\documentclass[sigconf]{acmart}
\usepackage{subfigure, algorithm, algpseudocode}

\copyrightyear{2019}
\acmYear{2019}
\acmConference[KDD '19]{KDD 2019: Workshop on Intelligent Information Feed}{August 04--08, 2019}{Anchorage, Alaska USA}
\begin{document}

\title{Understanding Video Content: Efficient Hero Detection and Recognition for the Game "Honor of Kings"}

\author{Wentao Yao}
\email{wentaoyao@tencent.com}
\affiliation{%
    \institution{Interactive Entertainment Group, Tencent Inc. (Shanghai)}
    \streetaddress{Tencent Building,Tower C,No.1081,Hongmei Road}
    \city{Xuhui District}
    \state{Shanghai}
    \country{China}
}

\author{Zixun Sun}
\email{zixunsun@tencent.com}
\affiliation{%
    \institution{Interactive Entertainment Group, Tencent Inc. (Shanghai)}
    \streetaddress{Tencent Building,Tower C,No.1081,Hongmei Road}
    \city{Xuhui District}
    \state{Shanghai}
    \country{China}
}

\author{Xiao Chen}
\email{evelynxchen@tencent.com}
\affiliation{%
    \institution{Interactive Entertainment Group, Tencent Inc. (Shanghai)}
    \streetaddress{Tencent Building,Tower C,No.1081,Hongmei Road}
    \city{Xuhui District}
    \state{Shanghai}
    \country{China}
}

\begin{abstract}
In order to understand content and automatically extract labels for videos of the game "Honor of Kings", it is necessary to detect and recognize characters (called "hero") together with their camps in the game video. In this paper, we propose an efficient two-stage algorithm to detect and recognize heros in game videos. First, we detect all heros in a video frame based on blood bar template-matching method, and classify them according to their camps (self/ friend/ enemy). Then we recognize the name of each hero using one or more deep convolution neural networks. Our method needs almost no work for labelling training and testing samples in the recognition stage. Experiments show its efficiency and accuracy in the task of hero detection and recognition in game videos.
\end{abstract}

    


\maketitle

\section{Introduction}
\label{sec:introduction}
As an essential part of a content-based recommendation system, content labelling plays an important role in content understanding and personalized recommendation. For the popular mobile game "Honor of Kings", there are numerous players that spend a lot of time playing this game or watching videos of this game everyday. When users are browsing game community, how to automatically recommend their favarite videos become an crucial problem for the operator of this game. An accurate recommendation will greatly motivate the users' interest and experience in this game. To understand and label the game videos, we must first detect and recognize the heros in the video.

There are usually two popular algorithm sets for neural network based object detection and recognition in images. One is called two-stage algorithms, which detect objects in image first and get bounding box for each detected object, then recognize each bounding box and get the category for each object. Typical CNN-based two-stage method includes R-CNN\cite{Girshick_2014_CVPR}, SPP Net\cite{He_2015_PAMI}, Fast R-CNN\cite{Girshick_2015_ICCV}, Faster R-CNN\cite{Ren_2015_NIPS} and Mask R-CNN\cite{He_2017_ICCV}, et al. The other algorithm set is called one-stage algorithms that directly detect and recognize objects in image in a single run, which typically includes SSD\cite{Liu_2016_ECCV} and YOLO(v1/2/3)\cite{Redmon_2016_CVPR}\cite{Redmon_2017_CVPR}\cite{Redmon_2018_arXiv}, et al.

We observe carefully videos of the game "Honor of Kings", and find some common characteristics for heros in videos of this game. Each hero, despite its camp, has a blood bar over it which indicates its life value. All blood bars have the same appearance (size and shape). The only difference between blood bars for different heros is the color, life value and level of the blood bar. This reveals an easy way to detect heros in video frames. Therefore, we adopt a two-stage method in this paper. 

In the first stage, all blood bars for heros in a video frame are detected based on a template-matching method and a list of bounding boxes for each video frame is obtained. In the second stage, we train a deep convolutional neural network to recognize each bounding box to get the name of the hero. 

The reason we adopt a two-stage method is that the blood bar for each hero has a fixed size and shape. Therefore it is expected to be efficiently and accurately detected with very few errors. Also, the recognition stage will benfit from the accuate detection result. Thus, we consider that our two-stage algorithm will outperform one-stage algorithms in this specific task. In addition, we have a set of game videos that only have the labels for the leading (self) hero. Therefore, the training and testing samples used for training classifiers could be automatically labelled using our detection algorithm by means of limiting the detection region around the center of video frames and the color of the blood bar to be green. On the contrary, for mainstream neural network based one-stage object detection algorithms, it is hard to manually label positions and names for all heros in video frames. If we only automatically label the leading heros which are located at the center of video frames from beginning to end, the trained neural network would tend to remember the position of the leading hero and have a bad detection result for other heros (freinds and enemies).

\section{Dataset}
\label{sec:dataset}
Because template-matching based algorithm will be used to detect blood bars in game videos, we expect all videos to have the same resolution and aspect ratio. However, we collect numerous videos of the game and find out various resolutions and aspect ratios. Fortunately, we discover the fact that blood bars in different videos have exactly the same size as long as the heights of these videos are same, which means the size of the blood bar is independent of the widths of videos.  Therefore, we normailize all videos to a standard height (720px) while keeping the aspect ratio unchanged. 

For the hero recognition task, it is necessary that we have plenty training and testing samples with labels. We have collected lots of game videos, including all 92 heros up to now, from some popular video websites. Each video has a label (name) of the leading hero. For each hero, we downloaded about 4-5 video clips with various appearances and skins of the hero (if possible). The leading hero detection algorithm will run on frames of each video to generate hero appearance training and testing samples. 

Besides the hero appearance classifier, we note that there will be a region for skills of the leading hero at the bottom right corner of the frame. For the same leading hero, the skill region is also exactly the same. We take advantage of the skill region and two additional convolutional neural network based classifiers are trained to improve recognition accuracy for the leading hero. One is based on the cropped whole skill region and the other is based on the detection result of the first skill circle. The position of skill region is not fixed and varies with the aspect ratio of the video. Therefore, the skill region detection algorithm should adapt the video's aspect ratio to extract the precise skill region. Also, a circle detection algorithm is used to detect the first skill circle based on the cropped skill region.

Therefore, for each hero, we have collected three types of samples: the appearance, first skill and skill region of the hero, which will be discussed in detail in Section \ref{sec:recognition}.

\section{Hero Detection in Video Frame}
\label{sec:detection}
\subsection{Blood Bar Template-Matching}
\begin{figure}[ht]
	\centering
    \subfigure[Blood bar template image]{
		\centering
		\includegraphics[width=0.45\linewidth]{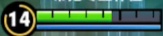}
		\label{fig:blood_bar_image}
	}
	\subfigure[Blood bar mask image]{
		\centering
		\includegraphics[width=0.45\linewidth]{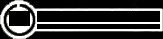}
		\label{fig:blood_bar_mask}
	}
    \caption{Blood bar template and mask images}
    \label{fig:blood_bar}
\end{figure}

The hero detection in game video is based on matching the blood bar over each hero with a predefined template. Since different blood bars have diverse life values, colors and levels, there must be a mask image indicating the region that will be used or not used for matching. Fig.\ref{fig:blood_bar} shows the blood bar template image and its corresponding mask image. Fig.\ref{fig:blood_bar_image} is the blood bar template used for matching and Fig.\ref{fig:blood_bar_mask} is the template mask where 1 (white) means pixels used for matching while 0 (black) means pixels not used for matching.

For a 3-channel input video frame, we first convert it to a grayscale image and perform template matching on the grayscale image. If the input image is not of normalized size (height = 720px), we first scale the input image to the normalized size. The matched image is a 32-bit floating image with each pixel indicating matching degree for input image and template at that position. We intend to detect all heros in a video frame, but the number of heros in one frame is uncertain. Therefore we could not apply a fixed threshold on the matched image, nor we could sort the matched values and pick up the first several values. To tackle this problem, we make observation on the original video frame and corresponding matched image, as in Fig.\ref{fig:origin_and_matched_image}:
\begin{figure}[ht]
	\centering
    \subfigure[Original video frame]{
		\centering
		\includegraphics[width=0.8\linewidth]{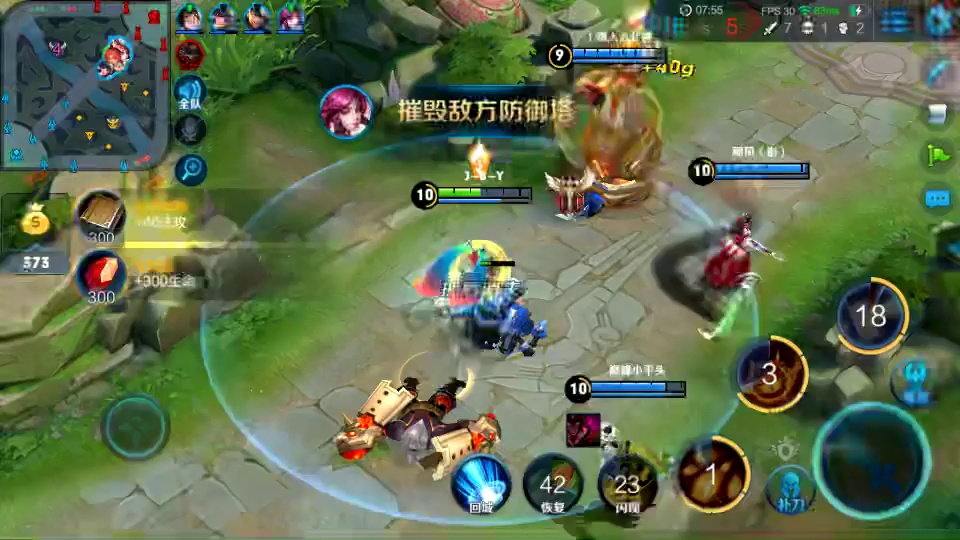}
		\label{fig:image_origin}
	}
	\subfigure[Corresponding matched image]{
		\centering
		\includegraphics[width=0.8\linewidth]{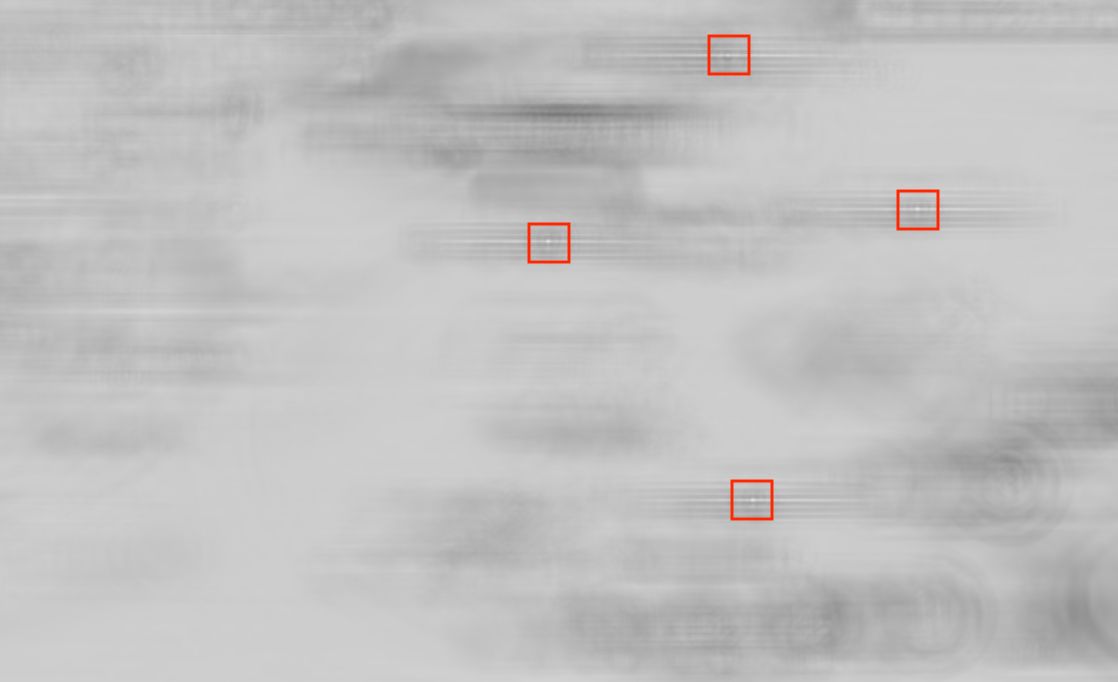}
		\label{fig:matched_image}
    }
    \subfigure[Local maximums of the matched image]{
        \centering
        \includegraphics[width=0.8\linewidth]{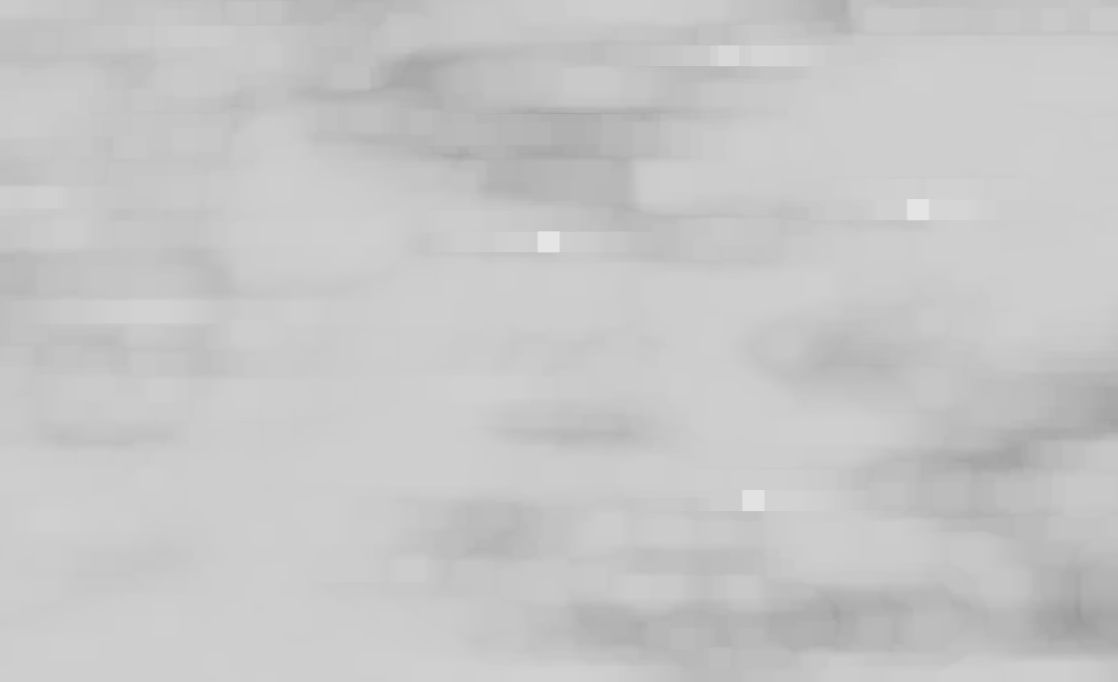}
        \label{fig:maximum_image}
    }
    \caption{Original video frame and its corresponding matched image}
    \label{fig:origin_and_matched_image}
\end{figure}

From Fig.\ref{fig:origin_and_matched_image}, we discover that for each blood bar, the matched image will have a local maximum value at the corresponding position. This means that there will be a pattern of serveral dark pixels surrounding a bright pixel in the matched image, as shown in the red box in Fig.\ref{fig:matched_image}. In Fig.\ref{fig:image_origin}, there exists four blood bars, and in Fig.\ref{fig:matched_image}, the above pattern appears at the corresponding positions. In contrast, the pattern is unobvious in other positions where no blood bar exists. This inspires us to detect blood bars by finding these local maximum patterns in matched image. We use a maximum filter with proper radius to filter the matched image. Fig.\ref{fig:maximum_image} is the image after maximum filtering. Obviously, the position of four maximum values corresponds to the four blood bars. 

We compare the matched image and the maximum image pixelwisely. An equal pixel value in the same position of the two images indicates a local maximum pixel. Typicallly, there will be hundreds of local maximum pixels in a matched image. Knowing the fact that there are at most 10 heros in one image, it is not necessary to process all these local maximum pixels. Instead, we sort all these local maximum pixels in descending order of their pixel values, and get the first 20 pixels for further processing. Experiments reveal that dropping out all the remaining maximum pixels will keep nearly all real blood bars, but significantly speeds up the detection.

After the top 20 local maximum pixels have been picked up, we design a function to compute a score for each local maximum pixel. The function is based on two considerations: local maximum pixel value and its contrast against surrounding pixels both contribute to a good template matching. We assume there are totally $n$ pixels in the maximum filter region. We denote local maximum pixel value as $v_0$, and the other pixel values in its filter region as $v_i (1\leq i\leq n), v_0\geq v_i$. We design the score function for each local maximum pixel as follows:
\begin{equation}
    score = \alpha*v_0 + \beta*\frac{1}{n}\sum_{i=1}^{n}(v_0-v_i)
\end{equation}
where $\alpha$ and $\beta$ are coefficients to balance the weights of two parts. The higher the score, the better the template matching result is. After we obtain scores for all 20 local maximum pixels, we sort these pixels in descending order by score (for further non-maximum suppression). Since we do not know the number of heros in the video frame, we still need a threshold to determine the number of heros. Applying a threshold in this stage is much better than in the template-matching stage. A fixed threshold works well for different frames in one video and also for frames from various videos.

\begin{figure}[ht]
	\centering
    \includegraphics[width=0.8\linewidth]{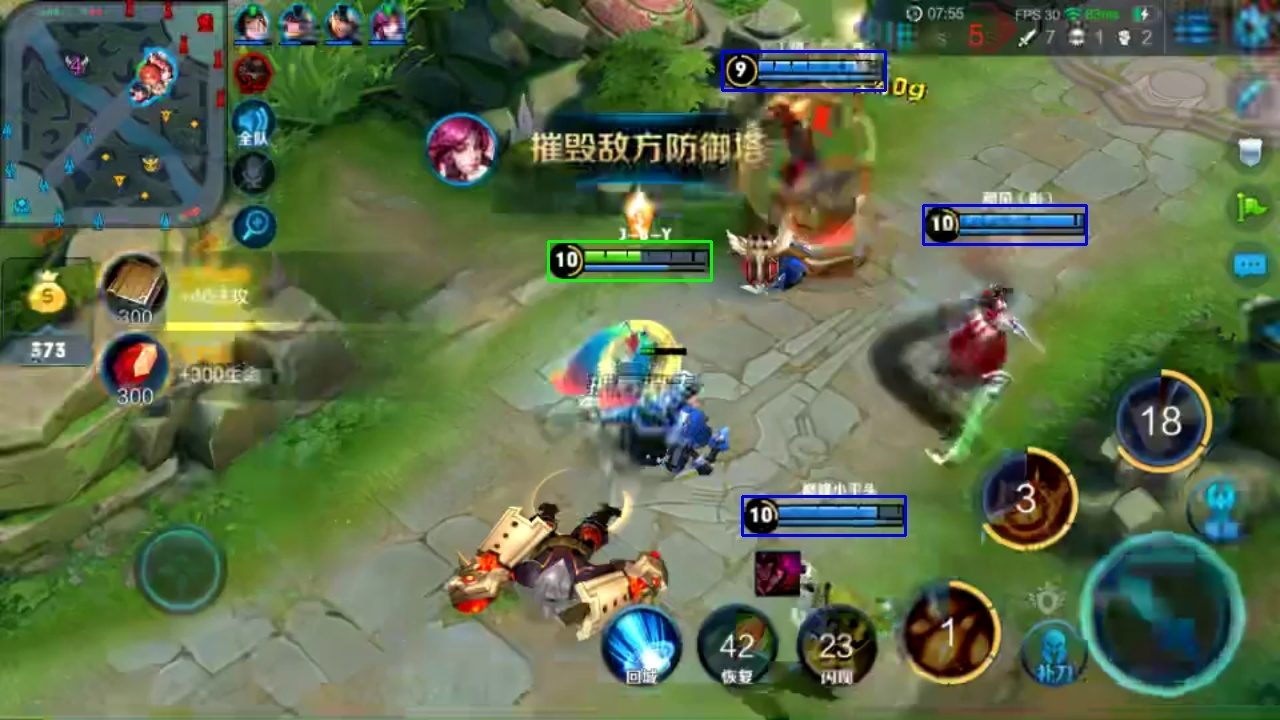}
    \caption{Blood bar detection result}
    \label{fig:image_show}
\end{figure}

The blood bar detection result for Fig.\ref{fig:image_origin} is shown in Fig.\ref{fig:image_show}. All blood bars for heros are correctly detected. 

\subsection{Non-Maximum Suppression}
\begin{figure}[ht]
    \centering
    \subfigure[Multiple detection results around the blood bar]{
        \centering
        \includegraphics[width=0.45\linewidth]{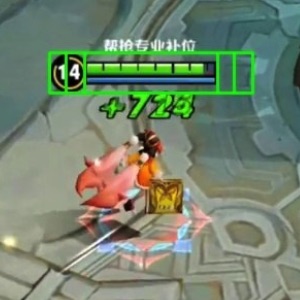}
        \label{fig:nms_before}
    }
    \subfigure[Detection result after non-maximum suppression]{
        \centering
        \includegraphics[width=0.45\linewidth]{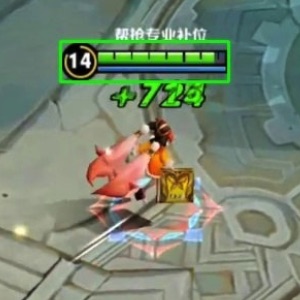}
        \label{fig:nms_after}
    }
    \caption{Detection results before and after non-maximum suppression}
    \label{fig:nms}
\end{figure}

In this game, the shape of hero's blood bar is nearly rectangle, which contains serveral long horizontal lines (shown in Fig.\ref{fig:blood_bar_image} and Fig.\ref{fig:blood_bar_mask}). Due to this fact, during the template-matching procedure, when the blood template moves horizontally around the real blood bar in image, the matching response will not decrease significantly (because most pixels on horizontal lines of template can still match the real blood bar pixels in image). As a result of this, there will be typically some adjecent detection results near the real blood bar, shown in Fig.\ref{fig:nms_before}.

To avoid multiple detection results for a same blood bar, non-maximum suppression is introduced. In the template-matching stage, we have already got the top 20 pixels with highest scores, and sort them in descending order by their scores. Each pixel is described using four properties $(x, y, score, is\_real\_detection)$. In non-maximum suppression stage, we design the suppression algorithm as in Algorithm \ref{alg:nms}, where $T_x$ and $T_y$ are thresholds for horizontal and vertical offset seperately. We set $T_x$ as half width of the template, and $T_y$ as 1 pixel. Fig.\ref{fig:nms_after} shows the detection result after non-maximum suppression, where all false detections are removed.
\begin{algorithm}
	\renewcommand{\algorithmicrequire}{\textbf{Input:}}
	\renewcommand{\algorithmicensure}{\textbf{Output:}}
	\caption{Non-maximum suppression}
	\begin{algorithmic}
		\Require Pixel set $S_i=\{P_1, P_2,\dots, P_n\}$, Let $P_i.is\_real\_detection\gets True$ for $1\leq i\leq n$
		\Ensure Pixel set $S_o\subseteq S_i$ that contains only real detections
        \For {$i \gets 2$ \textbf{to} $n$}
            \For {$j \gets 1$ \textbf{to} $i$}
                \If {$P_j.is\_real\_detection=True$ and $|P_j.y - P_i.y| < T_y$ and $|P_j.x - P_i.x| < T_x$}
                    \State $P_i.is\_real\_detection\gets False$
                \EndIf
            \EndFor
        \EndFor
        \State Output $S_o=\{P_i|P_i.is\_real\_detection=True\}$
    \end{algorithmic}
    \label{alg:nms}
\end{algorithm}

\subsection{Camp Classification for Heros}
By means of estimating the color of the blood bars, we can classify the heros into three camps: self (leading), friend and enemy. We adopt a simple algorithm to classify the blood bars using the average color of the left-most position in the blood bar. This is because the hero's life (blood) may be very low in some situations, which means most pixels in the blood bar is background that could not be used to estimate the camp of the hero. In addition, some false detections for blood bars can be eliminated in this stage. We denote the 3-channel average color of the left-most position in blood bar as $(c_r, c_g, c_b)$. The hero camp classfication rules are as follows:
\begin{itemize}
    \item For a color $i$ in (r, g, b), if $c_i > 100$ and $c_i > 1.5*c_j (j\neq i)$, then the blood bar's color is $i$ (green - self, blue - friend, red - enemy);
    \item If no $c_i$ matches the above rule, then:
    \begin{itemize}
        \item For every $i$ in (r, g, b), if $70\leq c_i\leq 100$ all satisfied, then the blood bar is almost empty (could not determine the hero's camp)
        \item Otherwise, it is a false detection for blood bar, remove the detection.
    \end{itemize}
\end{itemize}

The hero's camp classification is also shown in Fig.\ref{fig:image_show} and Fig.\ref{fig:failure}. The color of the bounding box for each blood bar indicates the corresponding camp of that hero. 

\begin{figure}[ht]
    \centering
    \subfigure[False detections for blood bars]{
        \includegraphics[width=0.8\linewidth]{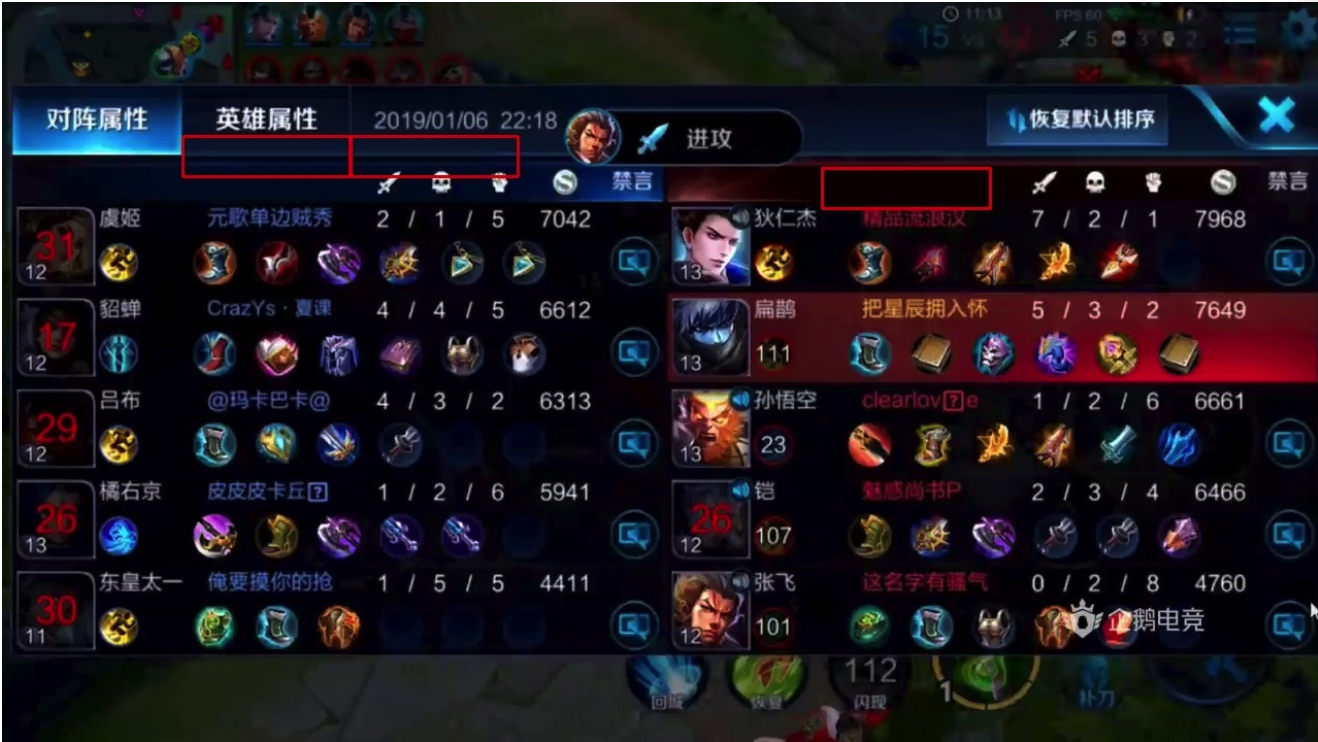}
        \label{fig:color_before}
    }
    \\
    \subfigure[Corresponding detection result after camp classification]{
        \includegraphics[width=0.8\linewidth]{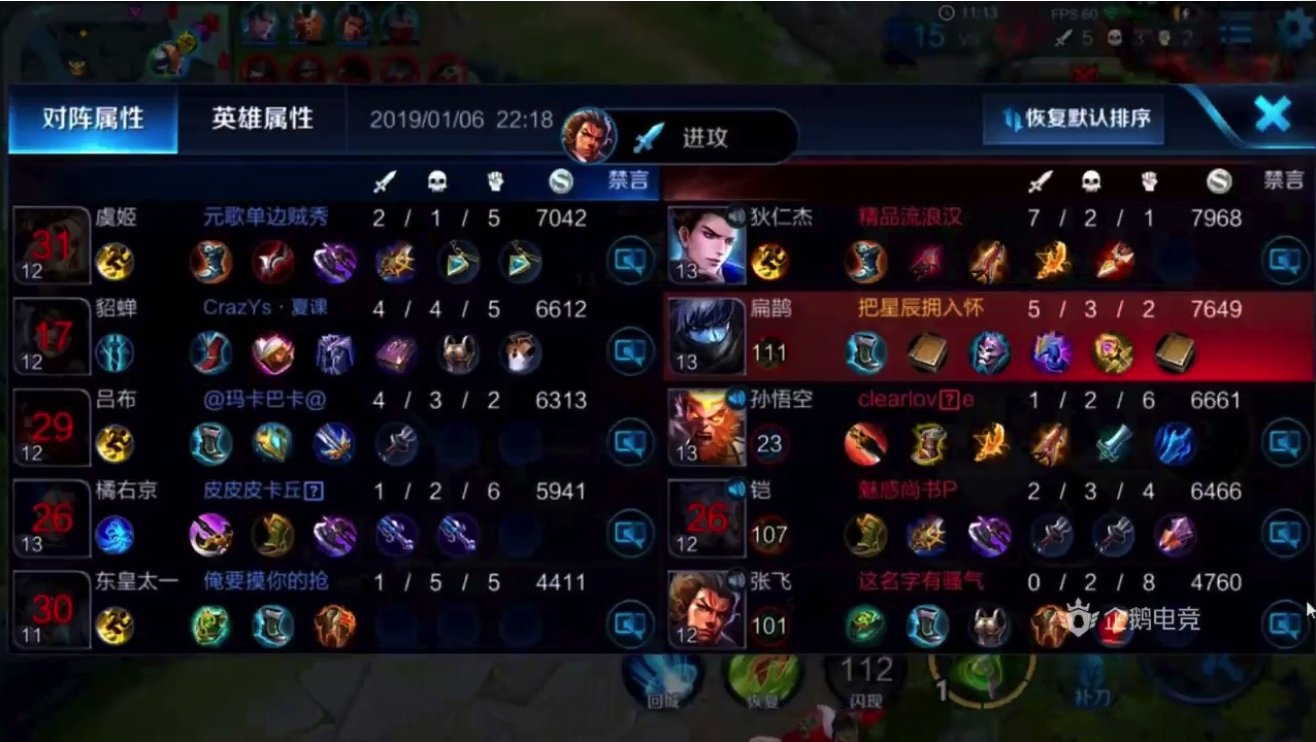}
        \label{fig:color_after}
    }
    \caption{Hero camp classification for item interface}
    \label{fig:camp_classification}
\end{figure}

Fig.\ref{fig:camp_classification} shows the camp classification for the shop interface during the game. Fig.\ref{fig:color_before} is the detection result for blood bars in this interface. We can see that there are some false detections due to many horizontal lines in this interface. Fig.\ref{fig:color_after} is the corresponding result after hero camp classification. We can find out that all false detections are successfully removed.

\section{Hero Recognition in Game Video}
\label{sec:recognition}
\subsection{Training and Testing Samples}
In order to recognize names of the heros in game video, we need to train several classifiers. As mentioned in Section \ref{sec:dataset}, we collect training and testing samples using the blood bar detection algorithm. 

For hero's appearance, we simply crop a fixed region under the detected blood bar for the leading hero. The image size for appearance is $163\times 163$.

For leading hero's skill region, as the position is not fixed due to different video aspect ratios, we make a position compensation according to the video's aspect ratio. In Eq.\ref{eq:skill_region}, variables ${x,y,w,h}$ with subscript $s$ are for skill region, and that with subscript $image$ are for image. $w_{norm}$ is the normalized image width in accordance with the normalized image height and standard aspect ratio (16:9). The coefficients are carefully adjusted to fit the most common aspect ratios. Although skill region cropping according to Eq.\ref{eq:skill_region} is still not very precise due to various aspect ratios and settings, it performs much better than cropping the region at a fixed position. Besides, to reduce false cropping of skill region during the non-game interface (Fig.\ref{fig:color_after}), the skill region is only cropped when the leading hero can be detected. The image size for skill region is $360\times 360$.
\begin{equation}
\begin{aligned}
    x_s &= 0.5 * w_{image} + 0.1875 * w_{norm}\\
    y_s &= 0.475 * h_{image}\\
    w_s &= 0.5 * h_{image}\\
    h_s &= 0.5 * h_{image}
\end{aligned}
\label{eq:skill_region}
\end{equation}

For hero's first skill, we run a circle detection algorithm at left bottom of the extracted skill region. If at least one circle can be detected, we crop the first skill region using the center of the largest circle and a fixed size. The image size for first skill region is $110\times 110$.

\begin{figure}[ht]
    \centering
    \subfigure[Appearance]{
        \begin{minipage}[b]{0.23\linewidth}
            \includegraphics[width=\linewidth]{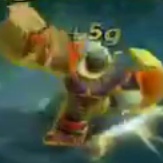}
            \\
            \includegraphics[width=\linewidth]{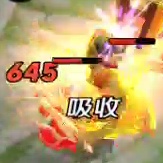}
        \end{minipage}
        \label{fig:lianpo_appearance}
    }
    \subfigure[Skill region]{
        \includegraphics[width=0.46\linewidth]{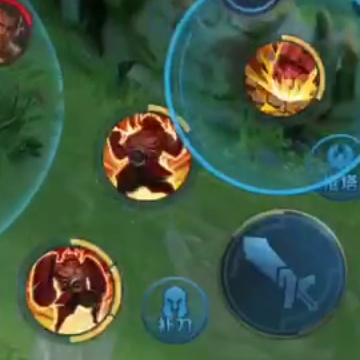}
        \label{fig:lianpo_skill_all}
    }
    \subfigure[First skill region]{
        \begin{minipage}[b]{0.23\linewidth}
            \includegraphics[width=\linewidth]{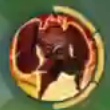}
            \\
            \includegraphics[width=\linewidth]{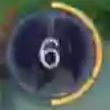}
        \end{minipage}
        \label{fig:lianpo_skill}
    }
    \caption{Typical training and testing samples}
    \label{fig:training_samples}
\end{figure}

Fig.\ref{fig:training_samples} shows typical training and testing samples used to train classifiers. Fig.\ref{fig:lianpo_appearance} shows images for hero's appearance. Note that heros may be dressed in difference skins, and may be occluded by other heros, text, digits or effect animation, which makes the appearance for the same hero diverse in the same or different videos. The two images in Fig.\ref{fig:lianpo_appearance} are of the same hero (Lianpo). Fig.\ref{fig:lianpo_skill_all} is the cropped skill region accoriding to Eq.\ref{eq:skill_region}. Fig.\ref{fig:lianpo_skill} displays the detected first skill region. Similarly, the images for skill and first skill region change over time and may be occluded by digits and effect animation. For example, the two images in Fig.\ref{fig:lianpo_skill} are of the same hero (Lianpo).

\subsection{Classifiers for Hero Recognition}
As mentioned above, we collect training and testing samples automatically using our detection algorithm. For each classifier, we have collected more than 100,000 samples. The numbers of training and testing samples are listed in Table \ref{tab:samples}.
\begin{table}
    \caption{Numbers of training and testing samples}
    \label{tab:samples}
    \begin{tabular}{lllll}
        \toprule
        Classifier & Sample Size & Total & Training & Testing\\
        \midrule
        Appearance & $163\times 163$ & 134,659 & 100,000 & 34,659\\
        Skill region & $360\times 360$ & 132,684 & 100,000 & 32,684\\
        First skill region & $110\times 110$ & 132,618 & 100,000 & 32,618\\
        \bottomrule
    \end{tabular}
\end{table}

We have trained our classifiers using three popular deep convolutional neural networks: Inception V3/V4 and Inception-ResNet V2\cite{Szegedy2016_AAAI}. The Inception network makes use of a parallel dimensionality reduction scheme to reduce the size of feature maps with fewer parameters to learn. For example, a $3\times 3$ convolution kernel can be replaced by a $3\times 1$ convolution kernel followed by a $1\times 3$ convolution kernel. The Inception-ResNet network is a combination of Inception and ResNet networks, which accelerates the convergence of the network by adding residual connections across layers. The performance for these classifiers will be shown in Section \ref{sec:experiment}.

\subsection{Whole Scheme for Hero Detection and Recognition}
\begin{figure}[ht]
    \centering
    \includegraphics[width=\linewidth]{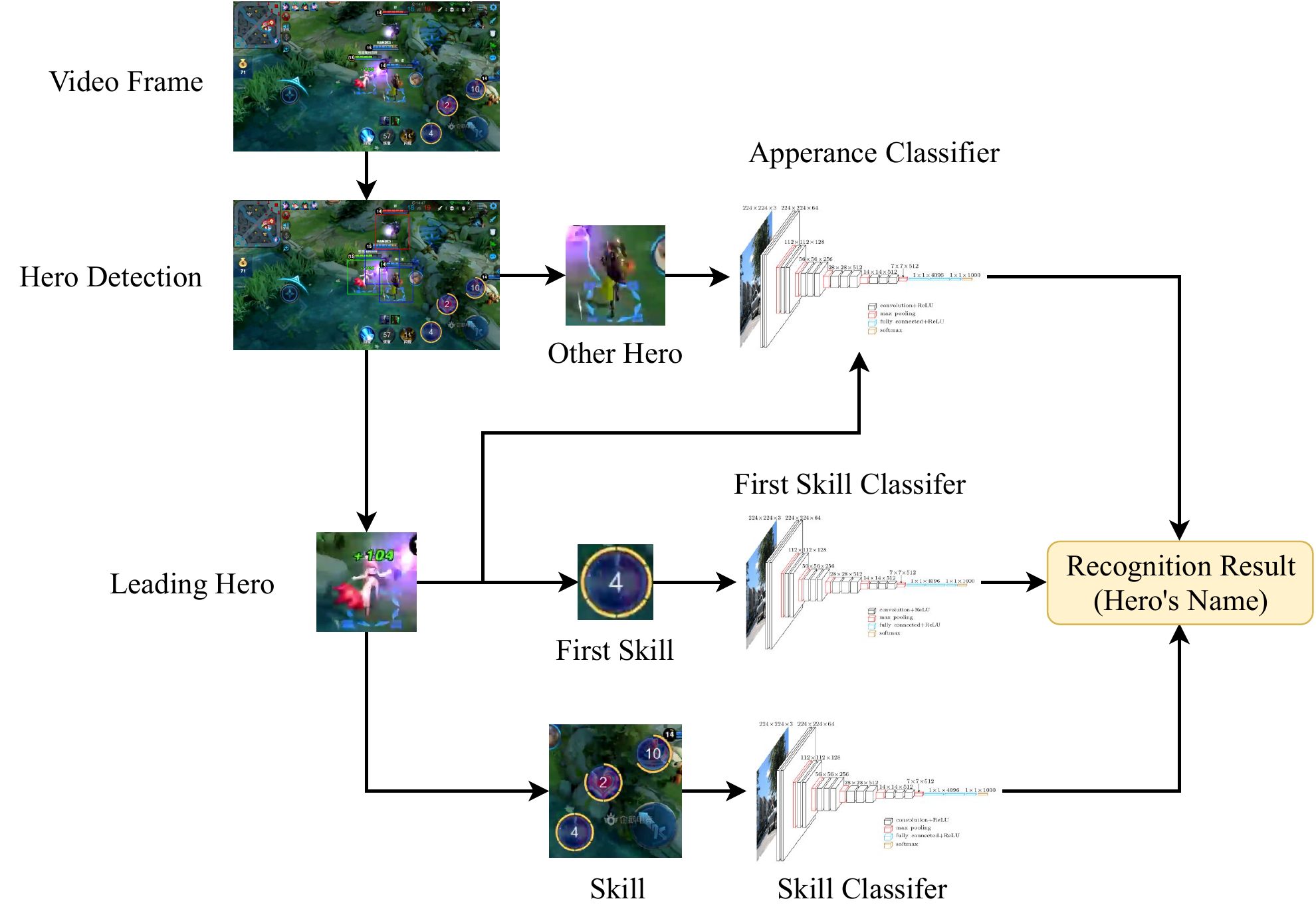}
    \caption{Whole scheme for hero detection and recognition}
    \label{fig:flowchart}
\end{figure}

Fig.\ref{fig:flowchart} is the whole sheme for our hero detection and recognition algorithm in a video frame. We run the hero detection algorithm on the input video frame to detect all heros in the image. For leading (self) hero, we send the cropped appearance, skill region and first skill region images into the three trained classifiers seperately. The final recognition result is based on the combination of labels and confidence scores of the three classifiers. For other heros, since no skill region is available, the recognition result is the label of the appearance classifier if the confidence score is above a threshold.

For a whole video composed of thousands frames, additional information can be utilized to get a more accurate recognition result rather than a single image. We run our detection and recognition algorithm on multiple frames sampled from the video. For each detected hero in each sampled frame, the name and confidence score of the hero will be obtained by the classifiers. The confidence scores for each hero are accumulated according to the name of the hero. After all the sampled frames are processed, we could obtain the final result by selecting the heros with highest accumulated confidence scores.

\section{Experimental Results}
\label{sec:experiment}
The input image size for Inception V3/V4 and Inception-ResNet V2 networks are all $299\times 299$. Therefore, for all three types of input images, we all scale them to $299\times 299$. We have collected videos for all 92 heros (up to now) in the game and extracted training and testing samples using our detection algorithm described in Section \ref{sec:detection}. We use average accuracy, marco-f1 and micro-f1 as evaluation criterior for three kinds of samples and three kinds of network models. The performance of the trained neural networks are listed in Table \ref{tab:performance}. All network models are trained on a Tesla M40 GPU and run on a GTX1060 GPU.
\begin{table*}
    \caption{The performance of the neural networks on testing set}
    \label{tab:performance}
    \begin{tabular}{llcccc}
        \toprule
        Image type & Network model & Average accuracy & Marco-F1 & Micro-F1 & Recognition Time(ms)\\
        \midrule
        Appearance & Inception V3 & \textbf{1.0000} & \textbf{0.9960} & \textbf{0.9962} & \textbf{4.1}\\
        Appearance & Inception V4 & 0.9998 & 0.9897 & 0.99 & 6.7\\
        Appearance & Inception-ResNet V2 & 0.9999 & 0.9943 & 0.9946 & 8.3\\
        \midrule
        Skill region & Inception V3 & \textbf{1.0000} & \textbf{1.0000} & \textbf{1.0000} & \textbf{4.7}\\
        Skill region & Inception V4 & 0.9989 & 0.9547 & 0.9497 & 7.0\\
        Skill region & Inception-ResNet V2 & \textbf{1.0000} & 0.9998 & 0.9998 & 8.8\\
        \midrule
        First skill region & Inception V3 & \textbf{1.0000} & \textbf{0.9986} & \textbf{0.9988} & \textbf{4.3}\\
        First skill region & Inception V4 & \textbf{1.0000} & 0.9985 & 0.9987 & 6.7\\
        First skill region & Inception-ResNet V2 & \textbf{1.0000} & 0.9976 & 0.9980 & 8.8\\
        \bottomrule
    \end{tabular}
\end{table*}

From Table \ref{tab:performance} we may find that the Inception V3 network outperforms the Inception V4 and Inception-ResNet V2 network for all types of images, because the Inception V3 network is enough to deal with these artificially synthesized images. Also, the Inception V3 network runs faster than the Inception V4 and Inception-ResNet V2 network on images because it is simpler in structure and has fewer parameters. The detection time for all heros in each frame is about 80ms. The whole process of detection and recognition for all heros in each frame, including leading hero's skill region and first skill region, is about 200ms for Incetpion V3 network, 280ms for Inception V4 network and 320ms for Inception-ResNet V2 network, respectively.

From the experiments, we find that Inception V3 network is better for the recognition tasks in game videos than the other two more complex networks, both in accuracy and recognition time. Therefore, we use the Inception V3 network as base network in our task.

\begin{figure}[ht]
    \centering
    \subfigure[Detection and recoginition result for YOLOv3 model]{
        \includegraphics[width=0.8\linewidth]{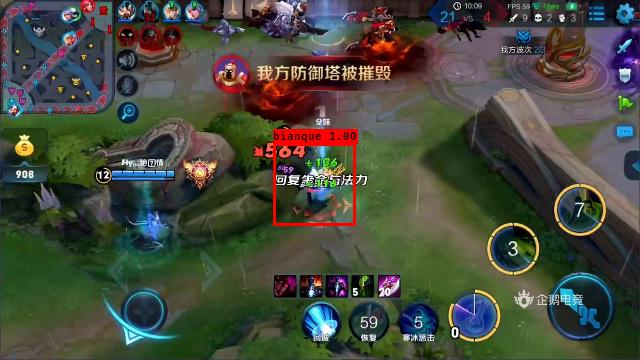}
        \label{fig:yolo_v3}
    }
    \\
    \subfigure[Detection and recognition result for our method]{
        \includegraphics[width=0.8\linewidth]{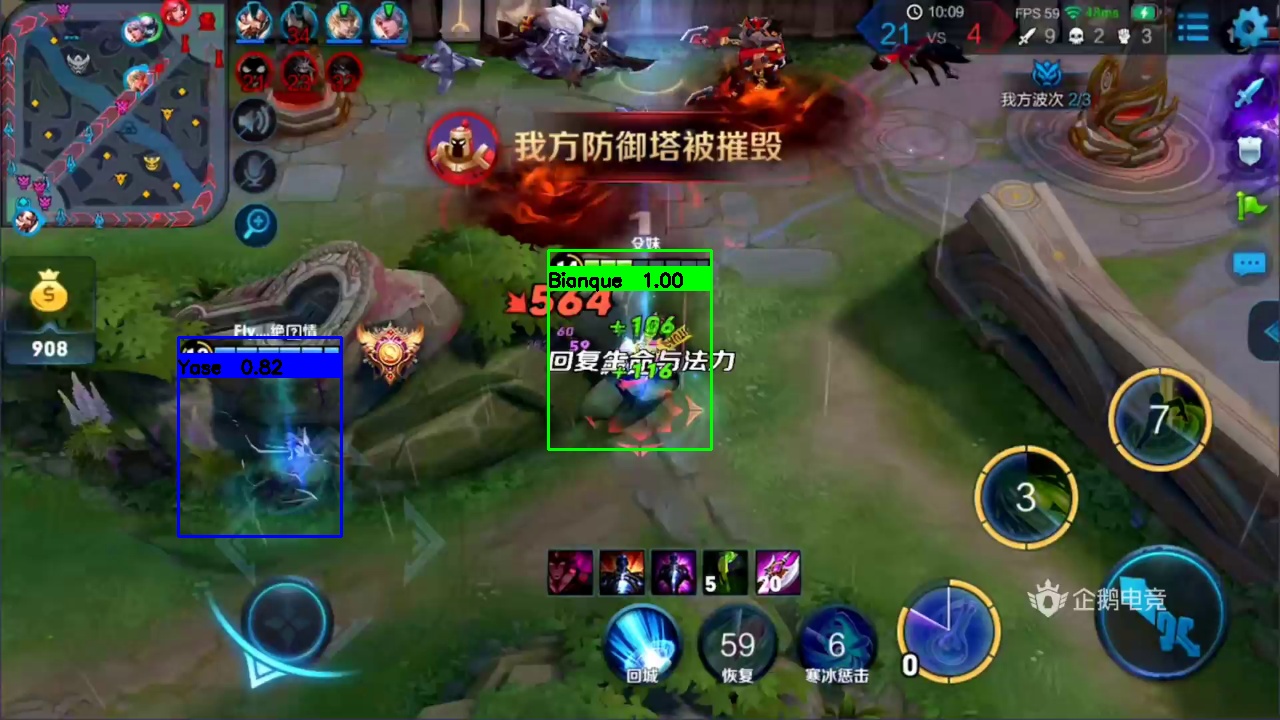}
        \label{fig:two_stage}
    }
    \caption{Comparison of YOLOv3 and our method}
    \label{fig:comparision}
\end{figure}

We also compared our method with YOLOv3, a popular one-stage object detection and recoginition model, as shown in Fig.\ref{fig:comparision}. As it is very difficult and tiring to label the position and name for each hero in a video, we only label the position and name of the leading hero in video frames, which is exactly the same way we collect samples for our method. Fig.\ref{fig:yolo_v3} shows the detection and recoginition result for YOLOv3 model. As analyzed above, the YOLOv3 model tends to detect only heros around the center of the image and miss the heros elsewhere, because the leading heros in training set are all around the center of the image. As a contrast, our method could detect and recognize all heros in the image, as shown in Fig.\ref{fig:two_stage}. 

\begin{figure}[ht]
    \centering
    \includegraphics[width=0.8\linewidth]{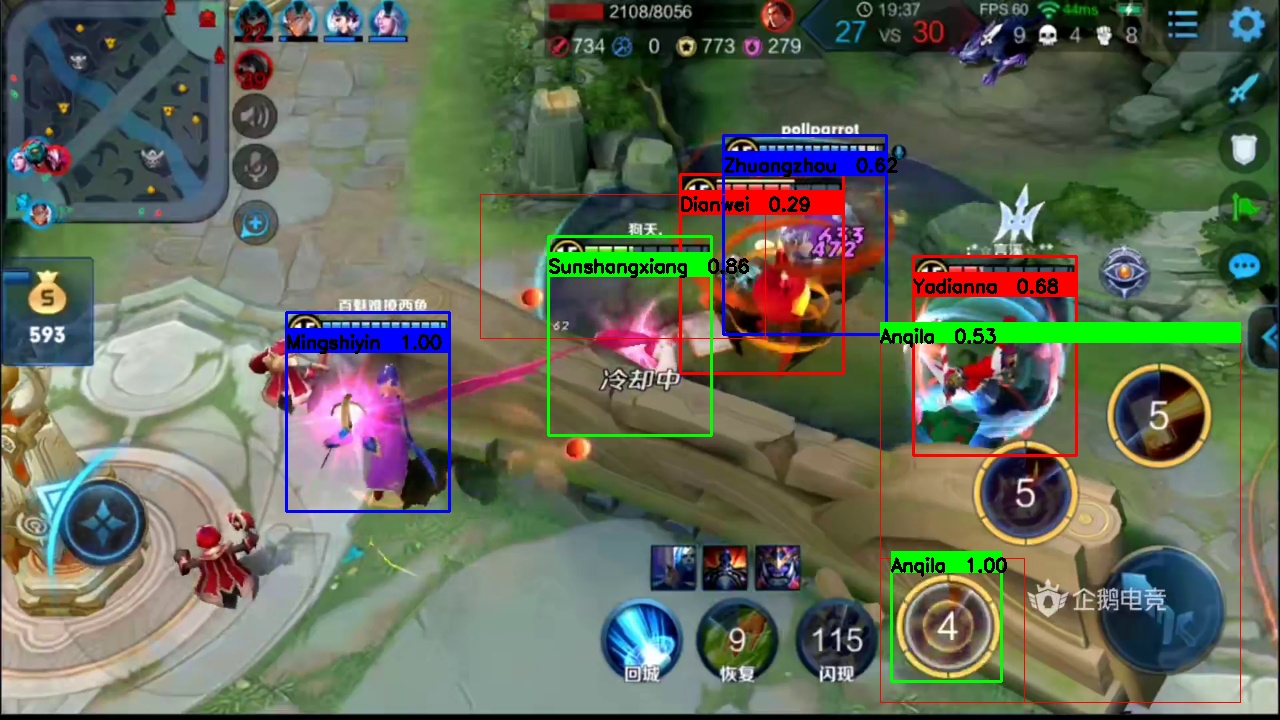}
    \caption{Typical recognition failure for hero's appearance}
    \label{fig:failure}
\end{figure}

Although our method performs well in experiments, there are still some situations that our recognition method will fail. The most typical recognition failure is a new skin for hero which did not appear in the training set. Also, the number of heros may increase with the update of the game. Thus, our model should be continuously updated to tackle these problems. Fig.\ref{fig:failure} shows a typical recognition failure for leading hero Anqila (mistakenly recognized as Sunshangxiang) when its skin did not appear in our training set. Although the recognition for hero's appearance fails, the skill and first skill region could still give correct results, which will help us recognize the leading hero correctly.

\section{Conclusions}
\label{sec:conclusions}
In this paper, we proposed an efficient and accurate hero detection and recognition algorithm for the game "Honor of Kings", which is helpful in game content understanding, video labelling and recommendation. We utilized a two-stage method to detect and recognize all heros in a video, including their names and camps. Our method outperforms popular one-stage methods, such as YOLO, with the same workload on labelling training and testing samples. Also, our method is efficient that could run at 5fps for $1280\times 720$ game videos. In the future, we will explore more information in game video content understanding, such as game video scene recognition and type classification. Besides, we will extend our work to other games such as "League of Legends", which will make our algorithm more general.

\bibliographystyle{ACM-Reference-Format}
\bibliography{ref}

\end{document}